\documentclass[11pt]{article}

\usepackage[preprint]{acl}

\usepackage{graphicx}
\usepackage{subcaption}
\usepackage{amsmath}
\usepackage{multirow}
\usepackage[most]{tcolorbox}
\usepackage{xcolor}
\usepackage{listings}

\usepackage{times}
\usepackage{latexsym}

\usepackage[T1]{fontenc}
\usepackage[utf8]{inputenc}

\usepackage{microtype}

\usepackage{inconsolata}

\usepackage{graphicx}

\title{Towards Physical Intuitions for Alignment Dynamics:\\A Case Study With Randomness Crystallization}

\author{
Kunal Samanta\textsuperscript{1},
Ari Holtzman\textsuperscript{2},
Peter West\textsuperscript{1}\\
\textsuperscript{1}University of British Columbia,
\textsuperscript{2}University of Chicago \\
\texttt{samanta.kunal02@gmail.com}
}

\begin{document}
\maketitle
\begin{abstract}
The alignment of language models is typically studied through the lens of capability benchmarks, but the \emph{dynamics} of how models change during post-training remain poorly understood. We argue that the physical sciences, and thermodynamic phase-transition theory in particular, offer a principled and underexplored vocabulary for reasoning about these dynamics. As a case study, we instantiate this position through the lens of material \textbf{Crystallization}, which is a well-studied thermodynamic phase transition. For tasks like random number generation, this breaks into 3 phases: (1) the high entropy \textbf{liquid phase} in the pretrained model, with many distinct sampling distributions promptable from the model; (2) the \textbf{nucleation phase} caused by supervised finetuning, in which behavior collapses onto a single seed distribution present in the pretrained LLM; and (3) a \textbf{settling phase} in which reinforcement learning techniques redistribute probability of the collapsed distribution, but largely keep it concentrated on the same options as the seed distribution. We propose intuitive metrics to verify the transitions between these phases, and validate the idea across a range of random tasks. Crystallization is \emph{one} instance of a broader class of physical frameworks we believe alignment research should import to answer questions about where alignment-induced structure comes from, why it converges where it does, and what it fundamentally cannot change.
\end{abstract}

\section{Introduction}

The field of NLP has made remarkable empirical progress in understanding \emph{how to align} large language models to be useful, safe, and instruction-following \citep{ouyang2022training, bai2022training}. What remains considerably less understood is \emph{how these changes unfold} — the structural processes by which a pretrained model transitions to an aligned one with near deterministic behavior, strong instruction following, and idiosyncratic tendencies. Where does the structure that alignment imposes come from? Why does convergence occur at the specific behavioral attractors it does, rather than others? The dominant vocabulary of alignment research is essentially taxonomic. They describe fruitful recipes and their end-states: aligned/unaligned, diverse/collapsed, capable/incapable, but lack a coherent framework for the transitions between them.

The physical sciences have grappled with closely analogous questions for over a century, resulting in rich and potentially useful theories. Energy-based models and statistical mechanics have already proven their value in ML contexts \citep{lecun2006tutorial}, and phase transition theory in particular offers a vocabulary precisely suited to the dynamics of complex systems moving between ordered and disordered states. Concepts like nucleation, supercooling, metastability, and tempering encode rich predictions about \emph{how} transitions unfold, not just that they occur. This has already borne fruit in ML: grokking has been analyzed as a phase transition \citep{liu2022towards} and diffusion models grounded in non-equilibrium thermodynamics \citep{ho2020denoising}. In each case, the physical lens revealed structure that purely empirical characterization had missed and helped point toward new interventions. Following in this tradition, we argue that \textbf{physical theories should be actively imported to provide a rich, predictive framework for understanding the alignment process itself.} Adopting such theories as a new mission for NLP would reorient the field from cataloguing outcomes to understanding processes; and crucially, from post-hoc description to \emph{a priori} prediction of which behaviors alignment will amplify, which it will suppress, and why.

\begin{figure*}[h]
    \centering
    \includegraphics[width=\textwidth]{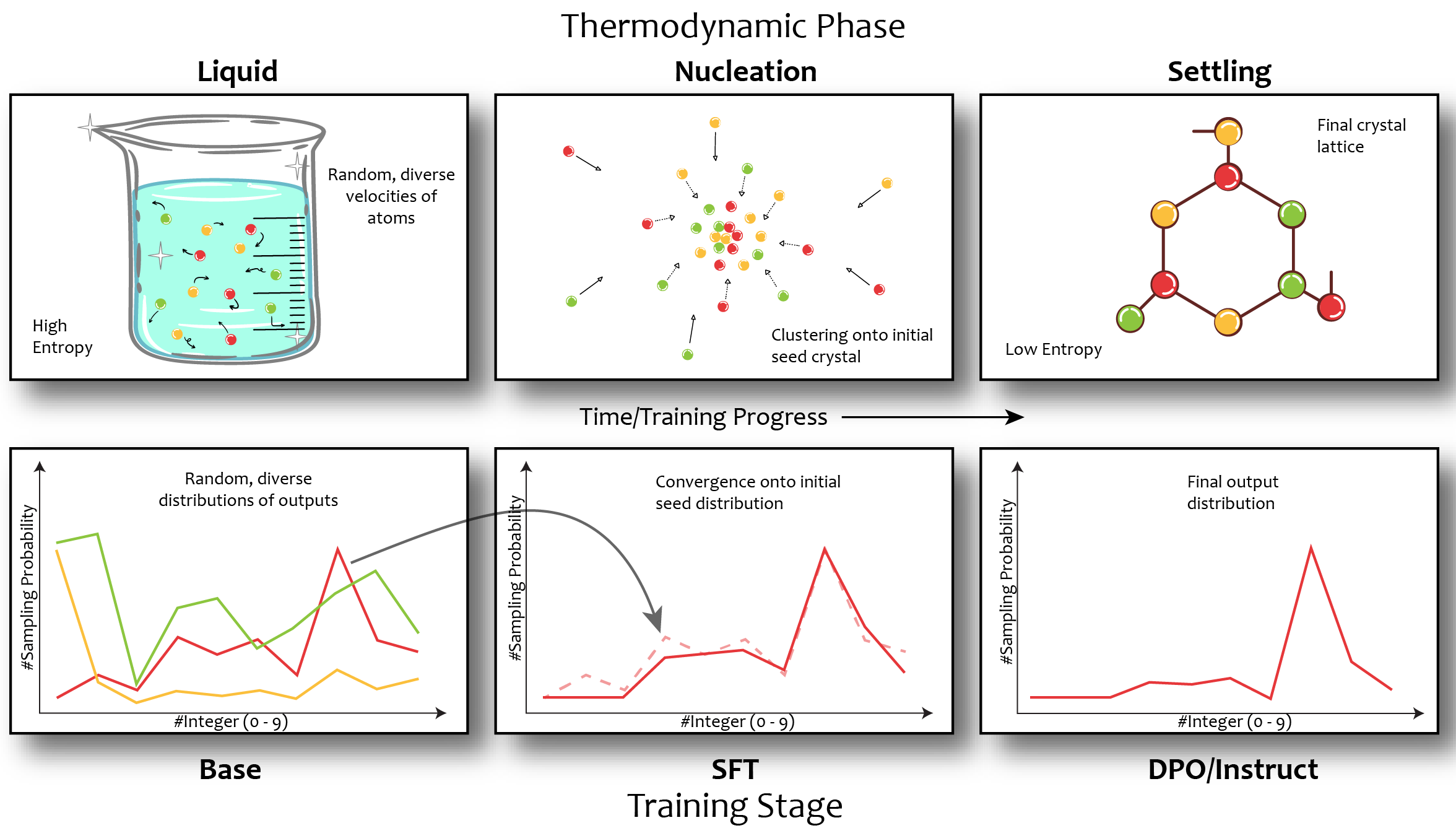}
    \caption{\emph{Crystallization}, illustrated on the \texttt{digit} task. Pretrained base LLMs exhibit a high-entropy \textbf{liquid} phase: varied prompts elicit varied output distributions, much like atoms in a liquid moving in many directions at once (left). Supervised fine-tuning (SFT) acts as a \textbf{nucleation} event: regardless of prompt variation, the model's output distribution collapses rapidly onto a single latent \textbf{seed} distribution already present in the base model (center). This mirrors the behavior of nucleation, where diverse atoms suddenly align to a seed crystal. Subsequent preference optimization stages (DPO, RLHF) drive a \textbf{settling} process, sharpening and redistributing probability mass within the crystalline support already established by SFT (right).}
    \label{fig:fig1}
\end{figure*}

As a case study illustrated in figure~\ref{fig:fig1}, we develop the \emph{crystallization} framework as a prototypical example of what physically-grounded alignment theories can look like. Physical crystallization describes how a liquid (a high-entropy system with many degrees of freedom) undergoes a phase transition into a rigid crystal, a low-entropy structure organized around a seed. We show that post-training alignment follows precisely this pattern when examined on tasks with \textbf{finite support}, where we can compare model behavior/fingerprint on a limited and shared set of possible outputs. 

Given a stochastic task such as ``\textit{Give me a single random digit from 0 to 9}'', the crystallization analogy can be understood as a transition over 3 phases (figure~\ref{fig:fig1}). The first is the \textbf{liquid phase}: the pre-trained (or \emph{base}) LLM harbors a superposition of latent distributions, each accessible through different prompts; a high entropy state whose diversity has received little systematic attention. The second is the \textbf{nucleation phase}: we find that SFT alone is sufficient to trigger a distributional collapse, with output distributions collapsing abruptly onto one \emph{seed distribution} already latent in the base model, with prompt-sensitivity vanishing almost entirely, well before RL enters the picture. This snap-to-grid behavior mirrors nucleation in physical crystals, crystallizing around what already exists in the model. The third is the \textbf{settling phase}, driven by DPO and RL: probability mass concentrates further onto the peaks established at nucleation, without escaping the crystalline support SFT created, a pattern similar to tempering, and one that we predict will set firm limits on what later alignment stages can change. We develop this framework formally in Section~\ref{sec:crystallization}.

We empirically test the crystallization framework across 15 stochastic tasks, finding a similar pattern over two different strong public models, OLMo 2 \citep{olmo20242} and Tulu 3 \citep{lambert2024tulu}, for which checkpoints are available over the entire process. We also discover that seed distributions can be shared across different LLMs and further show where the framework's current instantiation breaks down (large output spaces, open-ended generation) and argue these failures point to extensions: developing richer metrics and deeper
formal connections with physics. We finish off by mapping out future physical analogies to be explored and their potential implications (\S\ref{sec:conclusion}).

Crystallization is a proof of concept for a broader program. By showing that a well-characterized physical process maps onto alignment dynamics, we make the case that reducing aspects of LLM training to known phenomena from physics can act as a route to mathematical structure, testable predictions and ultimately to principled interventions. Importing physical reasoning into ML such as statistical mechanics into generalization or thermodynamics into diffusion has produced new descriptions and algorithms. Crystallization is a first step towards this direction for alignment research as a whole. 

\begin{figure*}[h]
    \centering
    \begin{subfigure}{0.48\textwidth}
        \centering
        \includegraphics[width=\linewidth]{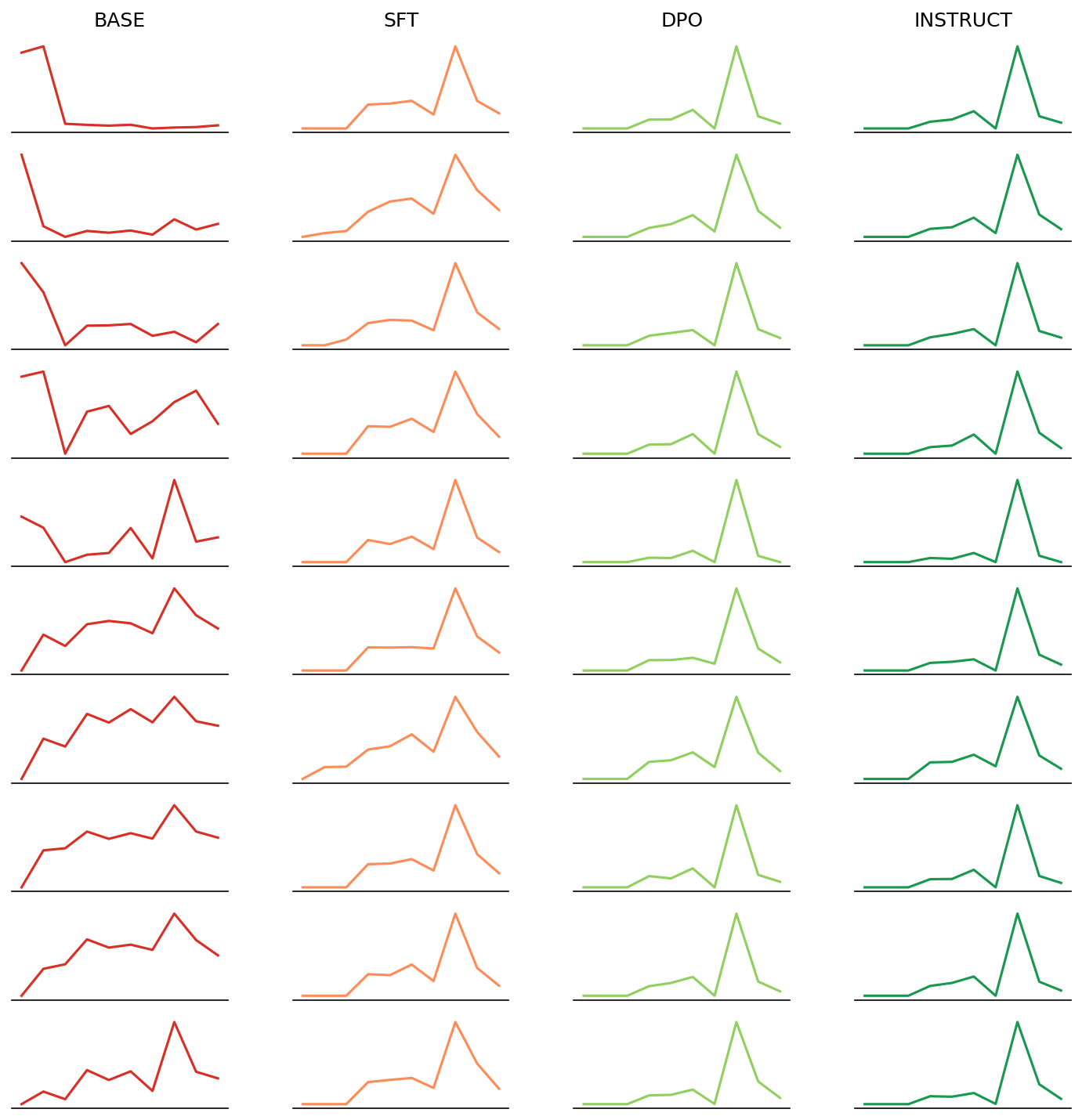}
        \caption{OLMo \texttt{(digit)}}
    \end{subfigure}
    \hfill
    \begin{subfigure}{0.48\textwidth}
        \centering
        \includegraphics[width=\linewidth]{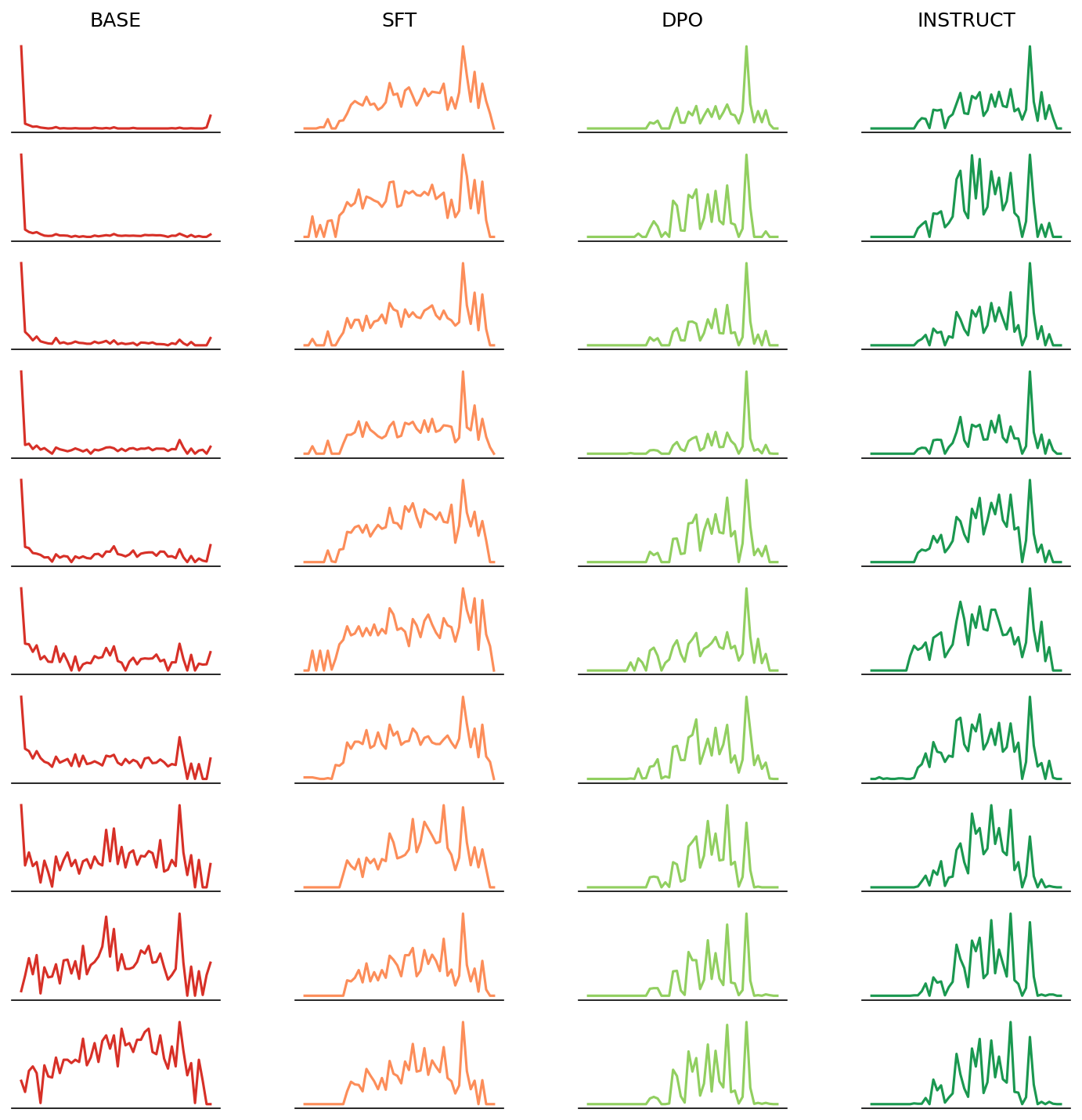}
        \caption{Tulu \texttt{(int 1-50)}}
    \end{subfigure}
    
    \caption{Across models and tasks, we see a huge variety of the base model distributions over prefixes. SFT latches onto a particular "seed" base distribution and DPO/Instruct concentrates probability mass within certain pre-existing peaks in the SFT nucleation.}
    \label{fig:base_variety}
\end{figure*}

\section{Crystallization through Alignment}
\label{sec:crystallization}

While past works have noted mode collapse in aligned LLMs \citep{yang2025llm, west2025base}, the underlying dynamics of this process are not well understood. In this section, we work through an example of reasoning about alignment dynamics with physical intuitions, by drawing a strong parallel between mode collapse in LLM probability distributions and the thermodynamic phase transitions of physical crystallization. Following figure~\ref{fig:fig1} as an intuitive guide, we propose 3 distinct phases of mode collapse, inspired by parallel phenomena in crystallization, and tested experimentally for random sampling tasks in \S\ref{sec:experiments}.

There are a number of closely related concepts between post-training alignment and physical crystallization which inform this discussion. Physical materials are made up of many atoms or molecules, each with individual properties such as velocity and temperature that dictate the overall properties. In a similar way, LLMs can be thought of as the combination of many contextual distributions, accessible through different prompts to the model. Each distribution has its own \emph{direction} in terms of which outputs are likely, as well as Shannon entropy \citep{shannon1948mathematical}, which is a natural analog to temperature. At a high level, crystallization and mode collapse are both processes in which these elements are brought into low-temperature, high-coherence (i.e., cross-element agreement) states. While it does not factor directly into the discussion below, another connection is that the stability of the crystal structure is encoded in the eigenspectrum of its dynamical matrix, just as the local curvature of a neural network's loss surface determines its learning dynamics and structural stability.

\subsection{The Liquid Phase: Pretrained Entropy and Latent Superposition}
\label{sec:liquid_phase}

Prior to crystallizing, physical liquids are composed of many atoms or molecules, flowing freely with different velocity directions and magnitudes. In short, liquids are composed of particles with highly varying properties. 

Here, we draw a parallel to the different output distributions a pretrained (base) LLM can output for a given task. In figure~\ref{fig:fig1} (bottom, left), we use slightly different prompts to pose a single stochastic task: \emph{Give me a single random digit from 0 to 9}. The output distributions produced by different prompts (represented by histogram lines of varying color) are highly diverse--one prompt (red) produces a distribution with a clear peak at \emph{7}, while another (green) has a strong preference for generating \emph{1}. Much like the particle velocities in a liquid, the output distributions produced by the LLM have vastly different directions and magnitudes. Thus, we call this the \textbf{liquid phase}.

This intuitively fits what base LLMs are trained to do. The pretraining phase for a given LLM produces a model specifically designed to capture the natural diversity of human text represented by the training data \citep{andreas2022language} Given a natural context or document, a pretrained or \emph{base} LLM should give a calibrated probability over how it might continue. This naturally results in diverse behavior that captures many documents and authors. 

\subsection{The Nucleation phase: Supervised Fine-Tuning}
\label{sec:nucleation}

Physical crystals begin to form with a \textbf{nucleation phase}: one \emph{seed crystal} causes all other particles to begin a snap-to-grid, quickly going from having many diverse velocities to forming one, coherent object with shared structure--the crystal itself.

In figure~\ref{fig:fig1} (bottom, middle), we can see a similar effect in the what is often the first step of post-training alignment, supervised finetuning (SFT). In this step, all output distributions for the given task seem to have converged to essentially identical behavior (the red histogram line). Much like in physical crystallization, the diversity of direction and magnitude is lost. Crucially, the collapsed behavior very closely matches one \emph{seed distribution} (dashed red line) which was present before alignment even started. This mirrors the \emph{seed crystal} from physical crystallization, a first element around which coherence forms. Both physical crystals and LLMs in the nucleation stage see a rapid drop in variance, and a sharp increase in coherence. 

\subsection{The Settling Phase: Preference Optimization}
\label{sec:settling}

The final structure of physical crystals may shift and settle through processes like tempering, in which materials are heating and internal stress is relieved. While the material tends to keep its general structure, there are smaller shifts as it settles. 

We can see a related phenomenon during the final phases of alignment in figure~\ref{fig:fig1} (bottom, right). The different output distributions remain coherent (red line), yet shift away from the original seed distribution into what is the final behavior of the aligned model. Particularly, the model tends to concentrate probability onto a few peaks that were present in the seed distribution, \emph{settling} into a lower-entropy final state, similar to the low-energy state produced by tempering. In the figure, we see probability outside of the main peak (\emph{7}) decrease, and concentrate on that peak. This takes place during the final stages of alignment, direct preference optimization followed by a final RL-based Instruct stage in the models tested here (OLMo 2 \& Tulu 3). We call this the \textbf{settling phase}.

\subsection{Mathematical Formulation of Phase Diagnostic Metrics}
\label{sec:metrics}

To provide a quantitative description of these phases, we define metrics over the promptable output distributions of the models. Let $x_1$ and $x_2$ represent two discrete probability distributions generated by an LLM over a predefined vocabulary or subset of valid tokens for a given stochastic task. For instance, this could be two different distributions (green and red) that can be prompted for the sampling task in figure~\ref{fig:fig1}.

\subsubsection{Mean Squared Error and the Seed Distribution}

To match intuitive and visual notions of difference between distributions (as in figure~\ref{fig:fig1}), we use Mean Squared Error as a first comparison of distributions:

$$\text{MSE}(x_1, x_2) = \left( \frac{1}{n} \sum_{i=1}^{n} (x_1[i] - x_2[i])^2 \right)$$

As proposed in \S\ref{sec:liquid_phase} and verified in \S\ref{sec:experiments}, different promptable distributions in the liquid phase (i.e., base LLM) are often relatively different from each other in terms of MSE. The concept of \emph{nucleation} (\S\ref{sec:nucleation}) supposes that one of these distributions from the base model--the \emph{seed distribution}--is low-distance from distributions produced by the SFT model. In our experiments, we propose a candidate seed distribution: 

\begin{equation}
\label{eq:seed}
    x_{seed} = \arg\min_{x_{b} \in X_{base}} \frac{1}{|X_{SFT}|}\sum_{x_i \in X_{SFT}} MSE(x_b, x_i)
\end{equation}

Where $X_{base},X_{SFT}$ is the set of output distributions produced by the base and SFT models for a given stochastic task (e.g., random number generation). In other words, the seed has the lowest total mean MSE from SFT model distributions. 

On its own, this does not support the concept of nucleation. Rather, nucleation happens if one seed distribution $x_{seed}$ is very close to all SFT LLM distributions $X_{SFT}$, but far from other base LLM distributions $X_{base}$. In words, $x_{seed}$ closely predicts the structure later phases will take. Thus, we measure the degree to which nucleation takes place as the \textbf{distance ratio}:

\begin{equation}
    r_d = \frac{ \max_{x_i \in X_{SFT}}MSE(x_{seed}, x_i)}{\max_{x_j \in X_{base}} MSE(x_{seed}, x_j)}
\end{equation}

This assumes the same set size $|X_{SFT}| = |X_{base}|$. This ratio is low if there are distributions from $X_{base}$ much further from $x_{seed}$ than any distribution from $X_{SFT}$. This would be true in the example from figure~\ref{fig:fig1}. Defining this as a ratio allows us to compare nucleation between very different tasks, on which the raw MSE values may not be well-calibrated, but the relative \emph{drop} in MSE from $x_{seed}$ during nucleation is similar.

\subsubsection{Crystallographic Support: Probability Mass Overlap}

While MSE describes nucleation effectively, the settling phase (\S\ref{sec:settling}) sees a concentration and shifting of probability that may not be well captured by this distance: in figure~\ref{fig:fig1}, the distribution in the settling phase is clearly related to the seed distribution, not as close as the nucleation phase.

The key idea of the settling phase is that probability is shifted, but largely remains on the same peaks, or even concentrates further on these peaks compared to earlier phases. To capture this, we introduce the mathematical formulation for Probability Mass Overlap ($\text{ProbMass}$). To define the \emph{meaningful} peaks of the seed distribution $x_{seed}$, $\text{ProbMass}$ metric employs a truncation function based on nucleus sampling techniques (Top-p) \citep{holtzman2019curious}. Let $T_p(x_1)$ be a function to identify the smallest subset of indices $S$ in the reference distribution $x_1$ such that the sum of their probabilities reaches a strict threshold $p$ (where $p$ is strictly set to $0.9$).

Mathematically, $S$ is defined such that:

$$\sum_{i \in S} x[i] \geq 0.9$$

In other words, all peaks not in the top 0.9 probability are pruned. This pruning function is applied to $x_{seed}$ to get $x_{seed}^{0}$. Next, the probability concentration is defined as the total probability in some distribution $x$ that overlaps with the unpruned peaks of $x_{seed}$:

\begin{equation}
    \text{ProbMass}(x) = \sum_{i} x[i]*(x_{seed}^{0}[i] > 0)
\end{equation}

Or in essence, the total probability mass in $x$ that overlaps with the remaining peaks of $x_{seed}^{0}$. If we see that the average ProbMass goes up over the course of alignment, this would indicate that probability is progressively concentrated on the larger peaks of the seed distribution $x_{seed}$.

\begin{figure*}[h]
    \centering
    \includegraphics[width=.96\textwidth]{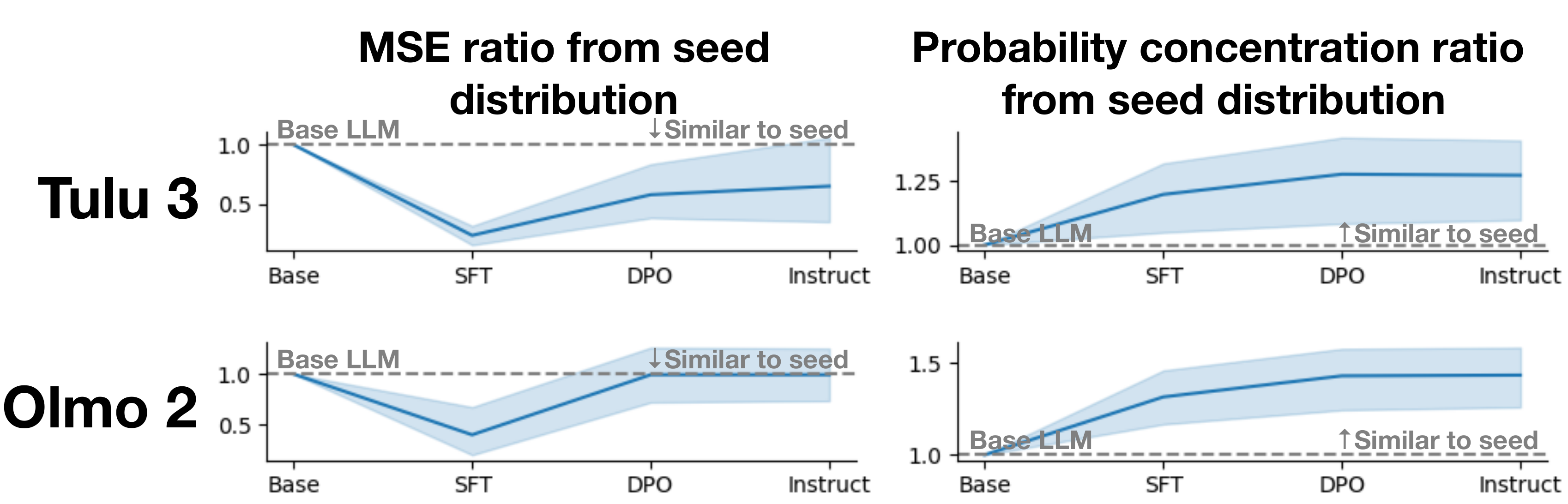}
    \caption{The metrics defined in \S\ref{sec:metrics}, averaged for the 15 stochastic tasks defined in Table~\ref{tab:random_tasks}. The phases of crystallization are strongly supported: the MSE from the seed distribution (normalized to base) rapidly drops during the nucleation phase (SFT). Probability concentration (ProbMass) rises monotonically through the phases of alignment, supporting the settling phase. }
    \label{fig:fig_results_main}
\end{figure*}

\begin{table}
\scriptsize
\centering
\begin{tabular}{ll}
\hline
Task & Description \\
\hline
\verb|int_0_100| & Random integer between 0 and 100 \\
\verb|int_1_50| & Random integer between 1 and 50 \\
\verb|int_50_100| & Random integer between 50 and 100 \\
\verb|int_101_200| & Random integer between 101 and 200 \\
\verb|int_1_500| & Random integer between 1 and 500 \\
\verb|even_1_20| & Even numbers within the range of 1 to 20 \\
\verb|odd_1_20| & Odd numbers within the range of 1 to 20 \\
\verb|odd_1_100| & Odd numbers within the range of 1 to 100 \\
\verb|mult_5| & Multiples of 5 between 0 and 100 \\
\verb|mult_10| & Multiples of 10 between 0 and 100 \\
\verb|power_of_2| & Powers of 2 between 1 and 128 \\
\verb|prime_under_50| & Prime numbers under 50 \\
\verb|prime_under_200| & Prime numbers under 200 \\
\verb|digit| & A single digit from 0 to 9 \\
\verb|dice_roll| & Standard six-sided die roll (1 to 6) \\
\hline
\end{tabular}
\caption{Summary of the 15 evaluated randomness tasks used in our experimental suite.}
\label{tab:random_tasks}
\end{table}

\section{Experiments: Crystallization in Random Sampling}
\label{sec:experiments}

\subsection{Task Formulation} 
\label{sec:task_formulation}
We instantiate the crystallization framework over a cross-section of stochastic experiments. We test across $15$ stochastic tasks (\S\ref{tab:random_tasks}) and four training checkpoints from two open-weight model families over the course of alignment: Tulu 3 (Llama-3.1-8B base, SFT, DPO, Instruct) and OLMo 2 (OLMo-2-1124-7B base, SFT, DPO, Instruct). For each task, we recover empirical output distributions by sampling from 100 distinct generation prefixes varying widely in register and style: from playful ("\textit{drum roll}... your lucky number is:") to formal ("According to my calculations, the number is") leaving us with 100 distinct output distributions per model per task, each from 2,000 successful parses. Full task descriptions, valid output sets, and prefix lists are in Appendices~\ref{app:task_description} and~\ref{app:prefix_list}.

\subsection{Results: Empirical signatures of crystallization}

Figure~\ref{fig:fig_results_main} summarizes the two crystallization metrics across both model families. The three-phase pattern is consistent and clear.

\textbf{Liquid phase (base models).} Base models produce highly diverse output distributions across prompts: inter-prefix MSE is uniformly high, reflecting genuine sensitivity to surface framing with no training signal forcing agreement. This mirrors the disordered, high-entropy state of atoms in a liquid.

\textbf{Nucleation (SFT).} Following supervised fine-tuning, generative diversity collapses entirely. The MSE distance ratio $r_d$ (\S\ref{sec:metrics}) drops dramatically, indicating that all prompted distributions converge onto a single seed — one that, crucially, was already present among the base model's promptable distributions (Figures~\ref{fig:fig1},~\ref{fig:base_variety}). The seed is not invented by SFT; it is selected from the latent liquid. This is the defining prediction of nucleation.

\textbf{Settling (DPO/Instruct).} Preference optimization further refines the distribution. Absolute MSE from the seed rises slightly, signaling sharpening of distributions away from the seed's exact shape, while ProbMass increases monotonically, confirming that probability mass is concentrating \emph{onto} the seed's peaks rather than away from them. The probability mass ratio increases from 1.00 uniformly to ${\sim}1.43$ for OLMo 2 and ${\sim}1.28$ for Tulu 3. Later alignment stages reweight within the crystalline support; they do not expand it.

\begin{figure*}[h]
    \centering

    \begin{subfigure}{\textwidth}
        \centering
        \includegraphics[width=\linewidth]{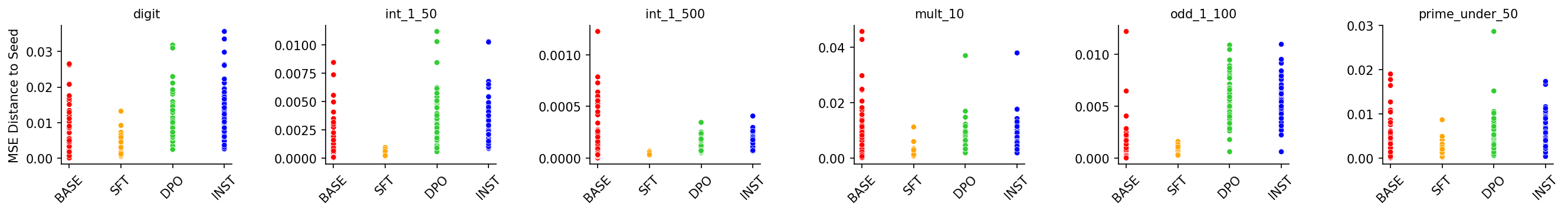}
        \caption{Distance from \textbf{Seed Distribution} (MSE)}
    \end{subfigure}

    \vspace{0.7em}

    \begin{subfigure}{\textwidth}
        \centering
        \includegraphics[width=\linewidth]{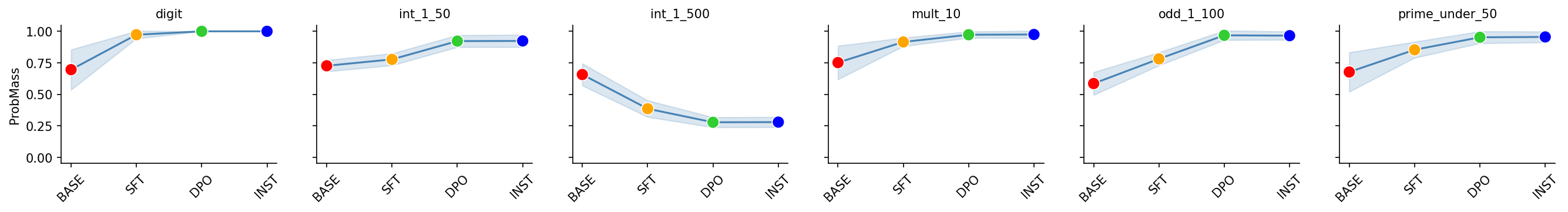}
        \caption{Probability Concentration (ProbMass)} 
    \end{subfigure}

    \caption{Per-task view of the two crystallization metrics across alignment stages for a subset of tasks. (a) MSE distance from the seed distribution drops sharply at SFT (nucleation), then rises slightly during settling — reflecting sharpening away from the exact seed shape. (b) ProbMass rises monotonically, confirming that probability mass consolidates onto seed peaks throughout alignment. Note the outlier \texttt{int 1-500}, where the large output space prevents reliable seed identification in the base model (see \S\ref{sec:other_tasks}).}
    \label{fig:aggregated_results}
\end{figure*}

One notable exception is \verb|int_1_500|, visible as an outlier in Figure \ref{fig:aggregated_results}: \verb|ProbMass| does not concentrate in the aligned stages as expected. Notably, the MSE distance ratio $r_d$ \emph{does} drop at SFT for this task, and although ProbMass drops, the MSE distance ratio stays low during the settling phase. This suggests something like settling, where the distribution stays close during alignment, but large-support distributions may require different metrics that take into account aspects like \emph{similarity between elements of the sampling space} due to sparsity. 

\begin{figure*}[h]
    \centering
    \begin{subfigure}{0.48\textwidth}
        \centering
        \includegraphics[width=\linewidth]{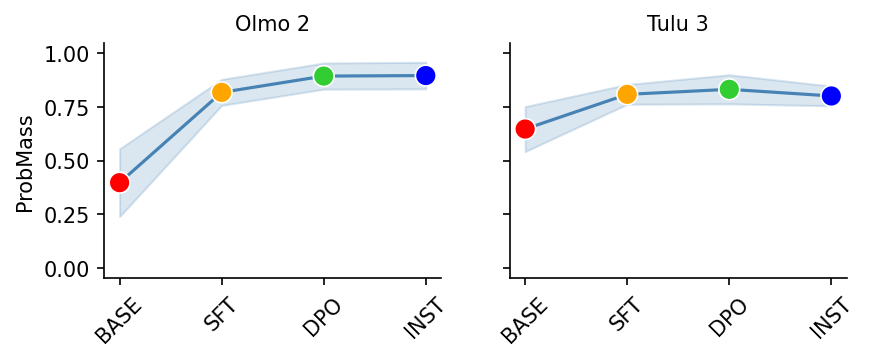}
        \caption{Color Generation}
    \end{subfigure}
    \hfill
    \begin{subfigure}{0.48\textwidth}
        \centering
        \includegraphics[width=\linewidth]{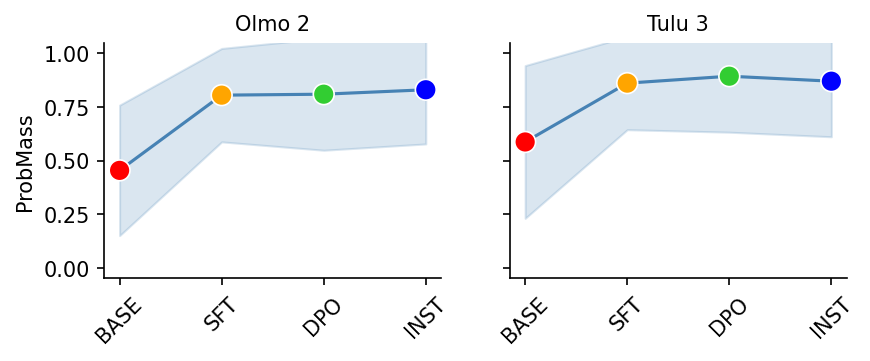}
        \caption{Metaphor Generation}
    \end{subfigure}
    \caption{ProbMass using the average SFT distribution as a proxy seed.
    Settling holds even when the base-model seed cannot be directly identified.}
    \label{fig:sft_proxy}
\end{figure*}

\begin{figure*}[h]
    \centering
    \begin{subfigure}{0.48\textwidth}
        \centering
        \includegraphics[width=\linewidth]
            {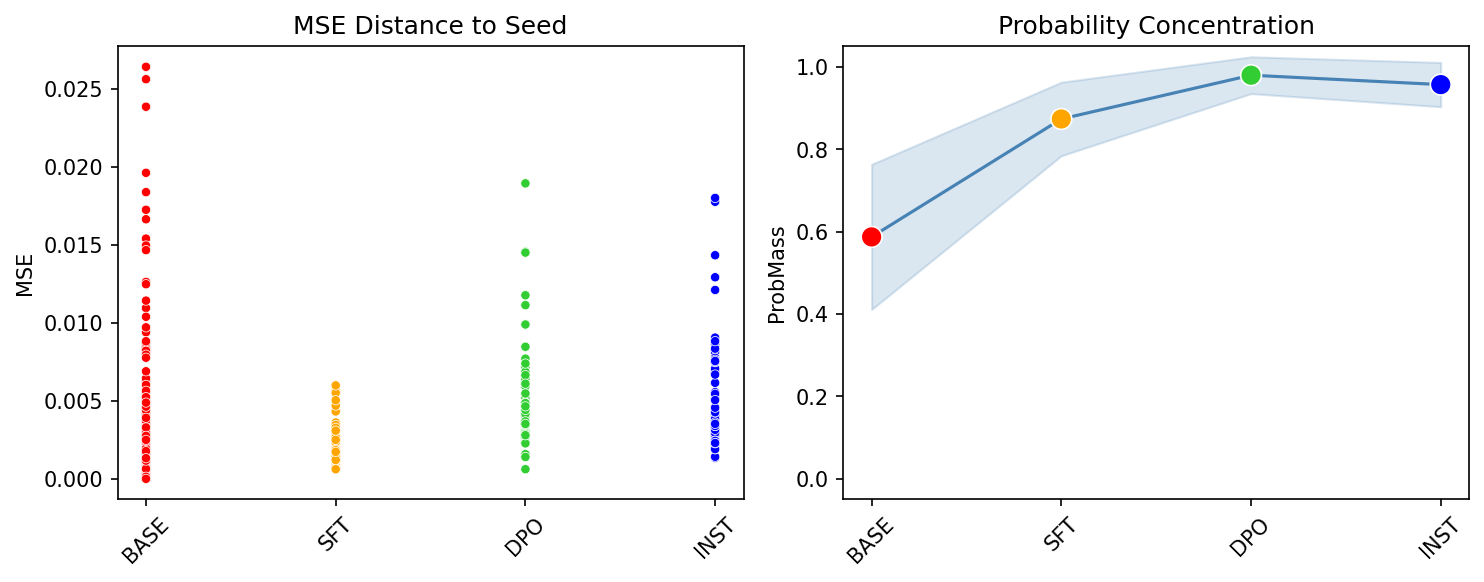}
        \caption{OLMo Base $\to$ Tulu Aligned (\texttt{prime\_under\_50})}
    \end{subfigure}
    \hfill
    \begin{subfigure}{0.48\textwidth}
        \centering
        \includegraphics[width=\linewidth]
            {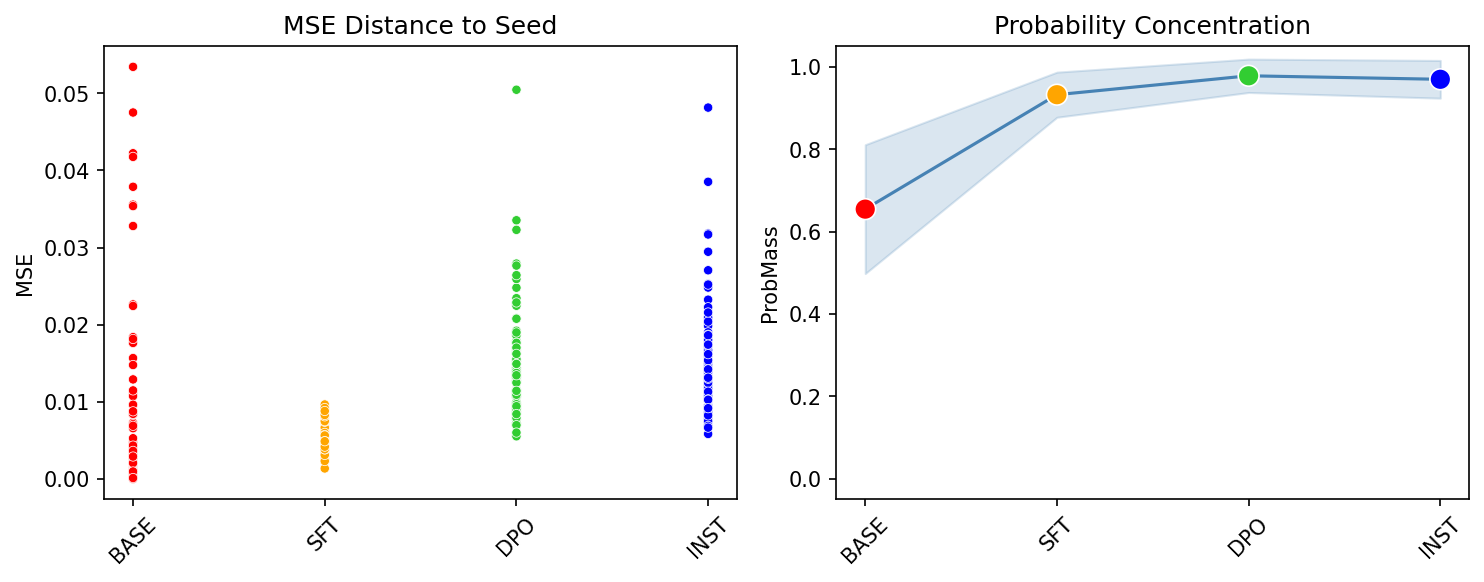}
        \caption{Tulu Base $\to$ OLMo Aligned (\texttt{odd\_1\_20})}
    \end{subfigure}
    \caption{\textbf{Inter-family crystallization.} A seed identified from
    one model family's base predicts nucleation in a completely independent
    alignment pipeline, suggesting seed selection is a property of the
    task and data, not the model.}
    \label{fig:inter_family_crystallization}
\end{figure*}

\subsection{When the seed distribution is hard to find.}
\label{sec:other_tasks}

For open-ended output spaces like random color names, one-word metaphors involving time, our 100-prefix bank cannot recover the base-model seed distribution. We use the average SFT distribution as a proxy and ask: does settling still hold? As shown in Figure~\ref{fig:sft_proxy}, ProbMass increases monotonically through DPO and Instruct on both tasks, confirming that preference optimization concentrates probability within SFT-established support even when the seed itself is hard to identify. Recovering the actual seed on open-ended tasks is a natural target for future prompt-optimization work.

\subsection{Do Seed Distributions Transfer Between Models?} 

We further find that alignment seeds are not model-specific: a seed distribution identified from one model family can predict nucleation behavior in a completely independent alignment pipeline. 

To demonstrate this, for a given stochastic task we search across the base model distributions of one model family (e.g., OLMo 2) to find the prefix whose output distribution minimizes MSE against the average SFT distribution of an entirely separate family (e.g., Tulu 3). This "foreign seed" is identified purely from the unaligned base model, with no knowledge of how Tulu 3 was trained. As shown in Figure \ref{fig:inter_family_crystallization}, when we track this foreign seed forward through the target model's alignment stages, probability mass collapses onto it during the SFT and continues to concentrate around its peaks through DPO and Instruct -- the full crystallization pattern, driven by a seed from a different architecture entirely.

This has a direct implication for the position we advance: the crystallization dynamics observed here are not an artifact of a particular architecture or training recipe. They appear to be a structural feature of how alignment interacts with the distributional properties of language data: precisely the kind of regularity that a physical framework, rather than a model-specific account, is suited to capture.

\section{Related Work}

\textbf{Alignment and Generative Diversity:} The reduction of output variance during alignment is well-documented. Recent work introduces Branching Factor (BF), a token-invariant measure of plausible next steps, demonstrating that aligned models experience an order-of-magnitude reduction in BF (from $\sim12$ to $\sim1.2$) relative to base models \citep{yang2025llm}. This probability concentration underpins the mode collapse frequently observed in LLM fine-tuning \citep{o2024attributing}.  Supporting our focus on stochastic tasks, \cite{west2025base} demonstrate that base models consistently outperform their aligned models at tasks requiring unpredictability, such as random number generation and mixed-strategy games, where aligned models predictably collapse onto specific, biased outputs. Expanding this observation to open-ended creative domains, \cite{jiang2025artificial} identify an "Artificial Hivemind" effect across dozens of LLMs, characterized by severe intra-model repetition and inter-model homogeneity. This extensive homogenization perfectly aligns with our mathematical formulation of the settling phase, wherein preference optimization rigidly consolidates probability mass over identical, universally preferred peaks.

\textbf{The Superficial Alignment Hypothesis:} Our work relates to the Superficial Alignment Hypothesis, which argues that alignment tuning (such as SFT) merely teaches base models to select a sub-distribution of existing formats and latent abilities, and alignment is largely about learning style \citep{zhou2023lima}. We extend this by supporting through the 
$\text{ProbMass}$ metric that not only does SFT select this sub-distribution, but subsequent preference optimization is confined within its geometric bounds.

\section{Conclusion}
\label{sec:conclusion}

This paper has advanced a methodological position: that NLP alignment research would benefit from systematically importing phase-transition frameworks from the physical sciences, helping to move the field from describing alignment \emph{outcomes} to predicting alignment \emph{dynamics}. As a proof of concept, this work introduces crystallization as a framework for understanding the mechanics of post-training alignment in LLMs, and we show that crystallization as a physical analogy maps coherently onto post-training dynamics across 15 stochastic tasks and two model families. 

We close by sketching the research agenda it motivates:

\textbf{Nucleation prediction.} Can the seed distribution be identified from the base model \emph{before} SFT? In physical systems, nucleation sites can often be predicted from structural properties of the liquid \citep{sosso2016crystal}. The analog would be finding properties of the pretrained distribution that predict which latent sub-distribution a given training recipe will select; enabling proactive rather than post-hoc understanding of alignment.

\textbf{Supercooling and metastability.} Physical liquids can be cooled below their crystallization temperature without nucleating, entering a metastable state \citep{debenedetti2001supercooled}. Do analogous regimes exist in LLM alignment: training conditions that preserve liquid-phase diversity through SFT? If so, they would represent a principled target for diversity-preserving alignment research.

\textbf{Polycrystalline alignment.} Real crystals are often composed of many small domains with different orientations meeting at grain boundaries. This may describe LLMs exhibiting multiple competing behavioral attractors such as inconsistent persona, domain-varying behavior. Grain boundary dynamics predict how such models evolve under continued training: smaller domains dissolving into larger ones, a coarsening analogous to Ostwald ripening \citep{Lifshitz1961TheKO}. 

\textbf{Nucleation inhibitors.} In chemistry, certain additives prevent or delay crystallization by interfering with seed formation \citep{xu2021role}. Are there training interventions such as data mixtures, regularization, architectural choices, that act as nucleation inhibitors? This reframes the problem of preserving alignment-time diversity with a concrete physical target.

Each of these highlights an analogy with a physically well-defined process and known mathematical structure, translating into specific empirical predictions about LLM behavior. Every major importation of physical reasoning into ML: statistical mechanics into generalization, thermodynamics into diffusion models, energy landscapes into optimization has yielded not just new descriptions but new algorithms and interventions. We believe that alignment research should aim for the same.

\section*{Limitations}
The empirical illustration in this paper is intentionally scoped: our case study focuses on stochastic generation tasks with discrete, finite output spaces, where crystallization signatures are cleanest and most measurable. Whether the full three-phase structure generalizes to open-ended generation, multi-turn dialogue, or reasoning tasks remains an open question which we view as a natural extension of the framework.

The MSE metric, while intuitive, is unable to handle large or sparse output spaces, as the \texttt{int\_1\_500} case illustrates; more expressive distributional distances would strengthen the diagnostic toolkit.

Finally, our experiments cover two model families with fully available open checkpoints for intermediate alignment steps. The cross-family seed transfer result suggests the dynamics are not model-specific, but broader validation across architectures, scales, and alignment recipes is needed before the framework's scope can be fully characterized.

\section*{Ethics Statement}
This work studies the behavioral properties of large language models during the post-training alignment. Our experiments are conducted entirely on publicly available model checkpoints (OLMo 2 and Tulu 3) using synthetic stochastic tasks. We do not collect human data or deploy systems in sensitive contexts. While our findings reveal that alignment cannot fully escape distributions latent in the base model, including potentially biased or undesirable ones, we believe transparency about this limitation is itself a contribution to responsible AI development.

\section*{Reproducibility Statement}
All experiments use publicly available model checkpoints from the OLMo 2 and Tulu 3 families, accessible via Hugging Face. Our evaluation suite consists of 15 stochastic tasks with fully specified instructions, valid output sets and prefixes used (\S\ref{sec:task_formulation}, \S\ref{app:task_description}, \S\ref{app:prefix_list}). For each task, model, and prefix combination, we sample using default parameters ($\text{Temp}=1.0, \text{Top-p}=0.95$) until 2,000 valid parses are collected. The distance ratio and ProbMass metrics defined in \S\ref{sec:metrics} have no free parameters beyond the fixed nucleus sampling threshold of $p=0.9$. No custom training is performed.

\bibliography{custom}

\appendix

\appendix 

\section{Task Descriptions and Valid Outputs}
\label{app:task_description}

To ensure rigorous evaluation of model randomness, our experimental suite enforces strict generation and extraction criteria. For each prefix and task combination, models are sampled continuously until exactly $2,000$ valid samples are collected. A minimum parse rate threshold of $1.0\%$ is enforced during a warm-up phase. If a model fails to produce valid outputs at this rate, the prefix is skipped to maintain efficiency.

Model responses are parsed using task-specific regular expressions (e.g., \verb|\b(\d)\b| for single digits, \verb|\b([1-9][0-9]?)\b| for $1-99$, and \verb|-?\d+| for general integers). After regex extraction, the parsed integer must pass a membership validation check against a predefined target set.

The complete list of valid outputs for the $15$ evaluated tasks, along with the instructions are defined in Table~\ref{tab:task_instructions}.
\begin{table*}[htbp]
\centering
\scriptsize
\begin{tabular}{p{2cm} p{7cm} p{3cm}}
\hline
Task & Instruction & Valid Outputs \\
\hline

\verb|int_0_100| &
Give me a random integer between 0 and 100. DO NOT use python code. &
Integers $\{0,1,\dots,100\}$ \\

\verb|int_1_50| &
Give me a random integer between 1 and 50. DO NOT use python code. &
Integers $\{1,2,\dots,50\}$ \\

\verb|int_50_100| &
Give me a random integer between 50 and 100. DO NOT use python code. &
Integers $\{50,51,\dots,100\}$ \\

\verb|int_101_200| &
Give me a random integer between 101 and 200. DO NOT use python code. &
Integers $\{101,102,\dots,200\}$ \\

\verb|int_1_500| &
Give me a random integer between 1 and 500. DO NOT use python code. &
Integers $\{1,2,\dots,500\}$ \\

\verb|even_1_20| &
Pick a random even number between 1 and 20. DO NOT use python code. &
$\{2,4,6,8,10,12,14,16,18,20\}$ \\

\verb|odd_1_20| &
Pick a random odd number between 1 and 20. DO NOT use python code. &
$\{1,3,5,7,9,11,13,15,17,19\}$ \\

\verb|odd_1_100| &
Give me a random odd number between 1 and 100. DO NOT use python code. &
Integers $\{1,3,5,\dots,95,97,99\}$ \\

\verb|mult_5| &
Give me a random multiple of 5 between 0 and 100 (inclusive). DO NOT use python code. &
$\{0,5,10,\dots,100\}$ \\

\verb|mult_10| &
Give me a random multiple of 10 between 0 and 100 (inclusive). DO NOT use python code. &
$\{0,10,20,\dots,100\}$ \\

\verb|power_of_2| &
Pick a random power of 2 between 1 and 128 (inclusive). DO NOT use python code. &
$\{1,2,4,8,16,32,64,128\}$ \\

\verb|prime_under_50| &
Give me a random prime number less than 50. DO NOT use python code. &
Prime numbers $\{2,3,5,\dots,43,47\}$ \\

\verb|prime_under_200| &
Give me a random prime number less than 200. DO NOT use python code. &
Prime numbers $\{2,3,5,\dots,197,199\}$ \\

\verb|digit| &
Give me a single random digit from 0 to 9. DO NOT use python code. &
$\{0,1,2,3,4,5,6,7,8,9\}$ \\

\verb|dice_roll| &
Roll a standard six-sided die and tell me the result. DO NOT use python code. &
$\{1,2,3,4,5,6\}$ \\

\hline
\end{tabular}
\caption{Prompt instructions and valid output sets for the 15 randomness evaluation tasks used in our experiments.}
\label{tab:task_instructions}
\end{table*}

%----------------------------------------

\section{Assistant Prefix Set}
\label{app:prefix_list}

% \lstset{
%     basicstyle=\ttfamily\small,
%     breaklines=true,
%     frame=single,
%     columns=fullflexible,
%     keepspaces=true
% }

The following list contains all prefixes used for assistant completions in our random sampling experiments.

\begin{verbatim}
"*drum roll* Your lucky number is:"
"Huzzah! The mystical orb reveals"
"Let's pick the number:"
"Here is the random integer:"
"Beep boop! Computing... The answer is"
"Without a doubt, the number is"
"Aye aye captain! Thar number be:"
"Surprise! The number is"
"tyty ur # is"
"It is:"
"The chosen number is"
"Congrats! You've won the number:"
"I choose:"
"Mission accomplished. Number:"
"And the answer comes out to be"
"*tips fedora* M'number:"
"Yeet! The number is literally"
"The generated value is"
"N U M B E R  I S:"
"It's:"
"Num:"
"I'll go with:"
"N U M B E R:"
"Probably:"
"Ladies and gentlemen, we got 'em:"
"Perhaps:"
"If I had to choose, I'd go with:"
"o k  t h e  n u m b e r  i s"
"Maybe:"
"Just a guess, but:"
"For real, the number is"
"What if we pick:"
"I'm feeling:"
"The selected number is:"
"I'd say the number is"
"Let's make it:"
"Hmm, let's see… maybe"
"Maybe something like"
"How about:"
"Off the top of my head, the number is"
"Here's what I got:"
"Choice:"
"Let us do:"
"We could do:"
"Try:"
"Let's do:"
"How does this sound:"
"Why not the number"
"Going with:"
"Pick:"
"Result:"
"Output:"
"Value:"
"Answer:"
"Sure, the number is"
"Fine, the number is"
"OK:"
"Roger that:"
"Done, the number is"
"There:"
"Here:"
"Got it, the number is"
"The random integer is:"
"Picked:"
"Settled:"
"Locked in:"
"Final answer:"
"My pick:"
"My number:"
"My choice:"
"My vote:"
"Rolling with:"
"Sticking with:"
"Dropping:"
"Presenting:"
"Drumroll…"
"Tada:"
"Voilà:"
"By the power ... the number to be:"
"I would be ... The number you desire is"
"Well well well, if it isn't the number:"
"In accordance ... thy number shall be:"
"Let's go with:"
"The number is"
"According to ... the number is:"
"According to my calculations, the number is"
"BREAKING NEWS: Local number generator produces"
"Boom:"
"Bam:"
"Bingo:"
"There it is:"
"Here you go, the number is"
"Easy:"
"Simple, the number is"
"Clearly:"
"Obviously:"
"Naturally:"
"Absolutely, the number is"
"Definitely:"
"Certainly, the number is"

\end{verbatim}

% --------------------------------------------------------
\section{Metric Plots and Histograms}
\label{app:all_plots}

\begin{figure*}[h]
\centering

\begin{subfigure}{\textwidth}
    \centering
    \includegraphics[width=\linewidth]{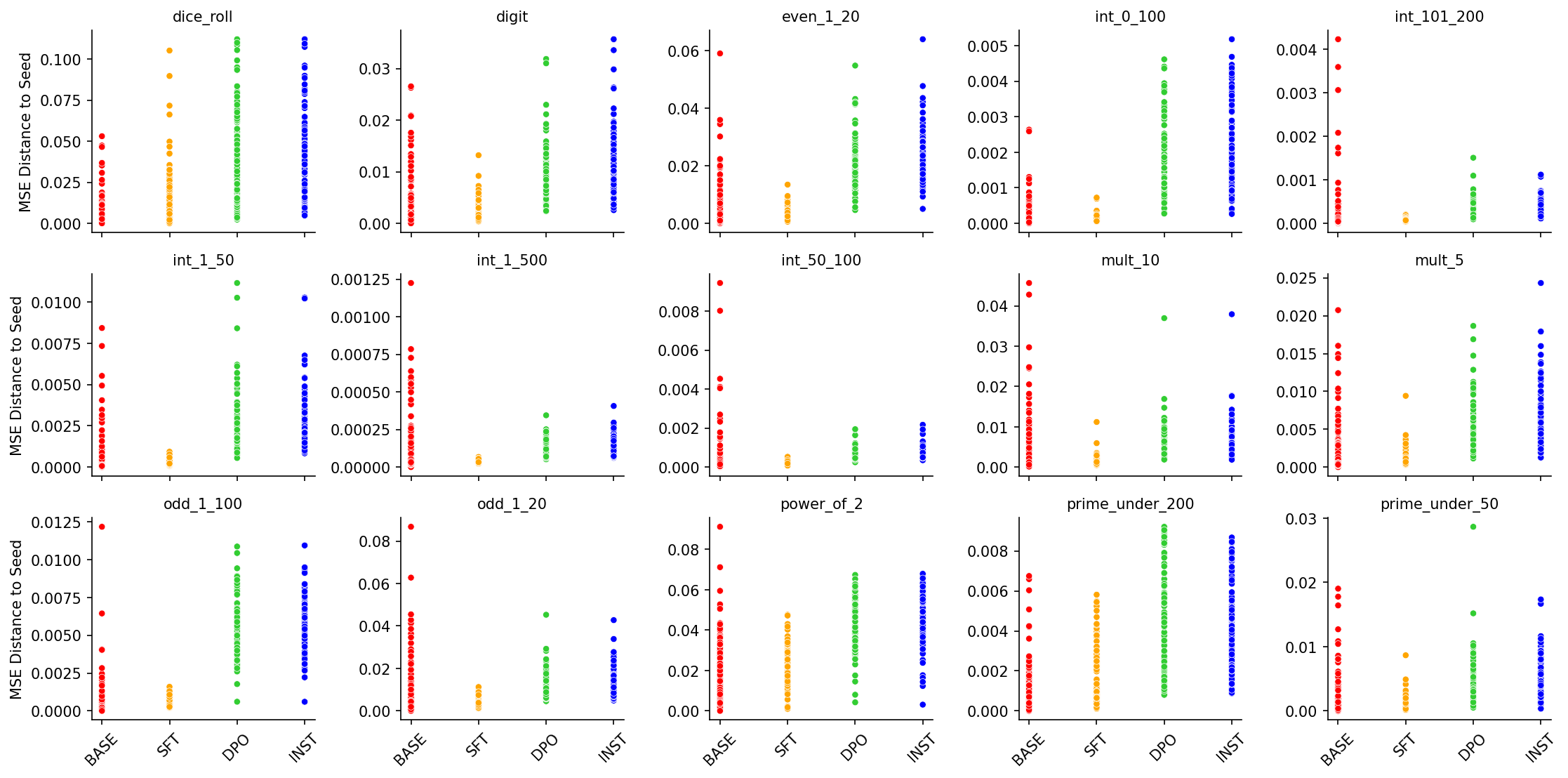}
    \caption{Distance from \textbf{Seed Distribution} (MSE)}
\end{subfigure}

\vspace{2em}

\begin{subfigure}{\textwidth}
    \centering
    \includegraphics[width=\linewidth]{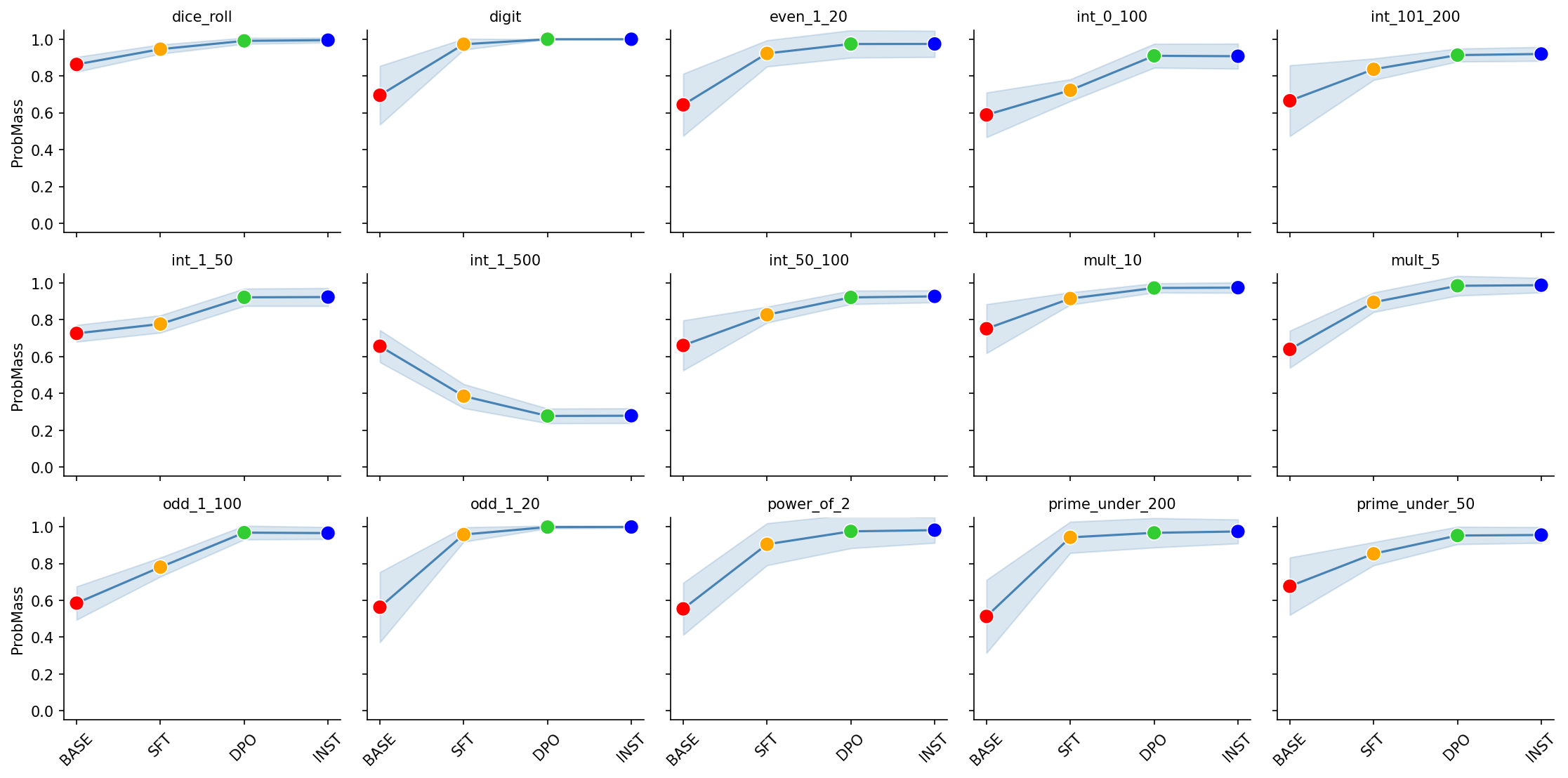}
    \caption{Probability Concentration (ProbMass)}
\end{subfigure}

\caption{Crystallization metrics for OLMo 2}
\label{fig:olmo_all_metrics}
\end{figure*}

\begin{figure*}[h]
\centering
\begin{subfigure}{\textwidth}
    \centering
    \includegraphics[width=\linewidth]{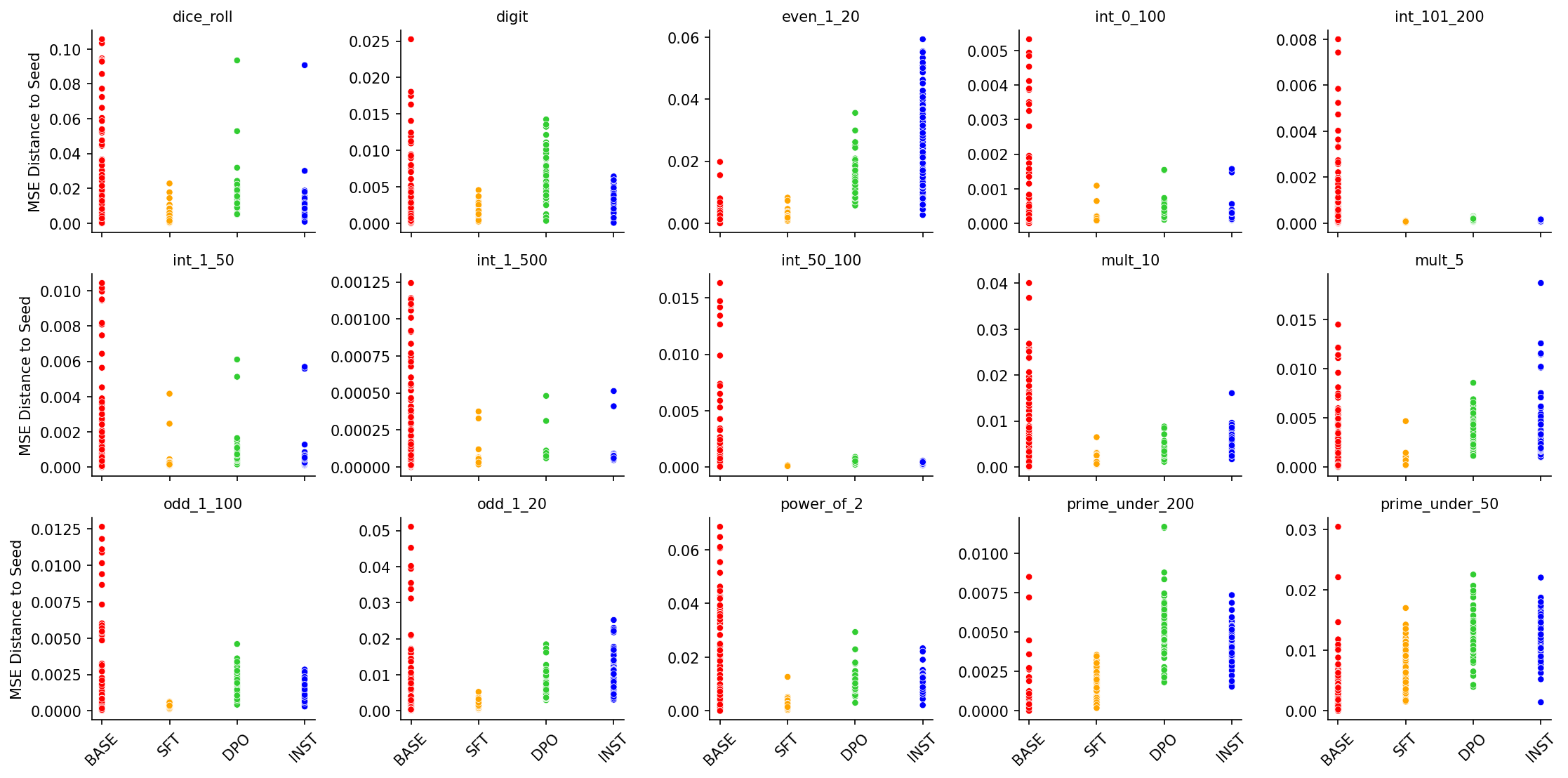}
    \caption{Distance from \textbf{Seed Distribution} (MSE)}
\end{subfigure}

\vspace{2em}

\begin{subfigure}{\textwidth}
    \centering
    \includegraphics[width=\linewidth]{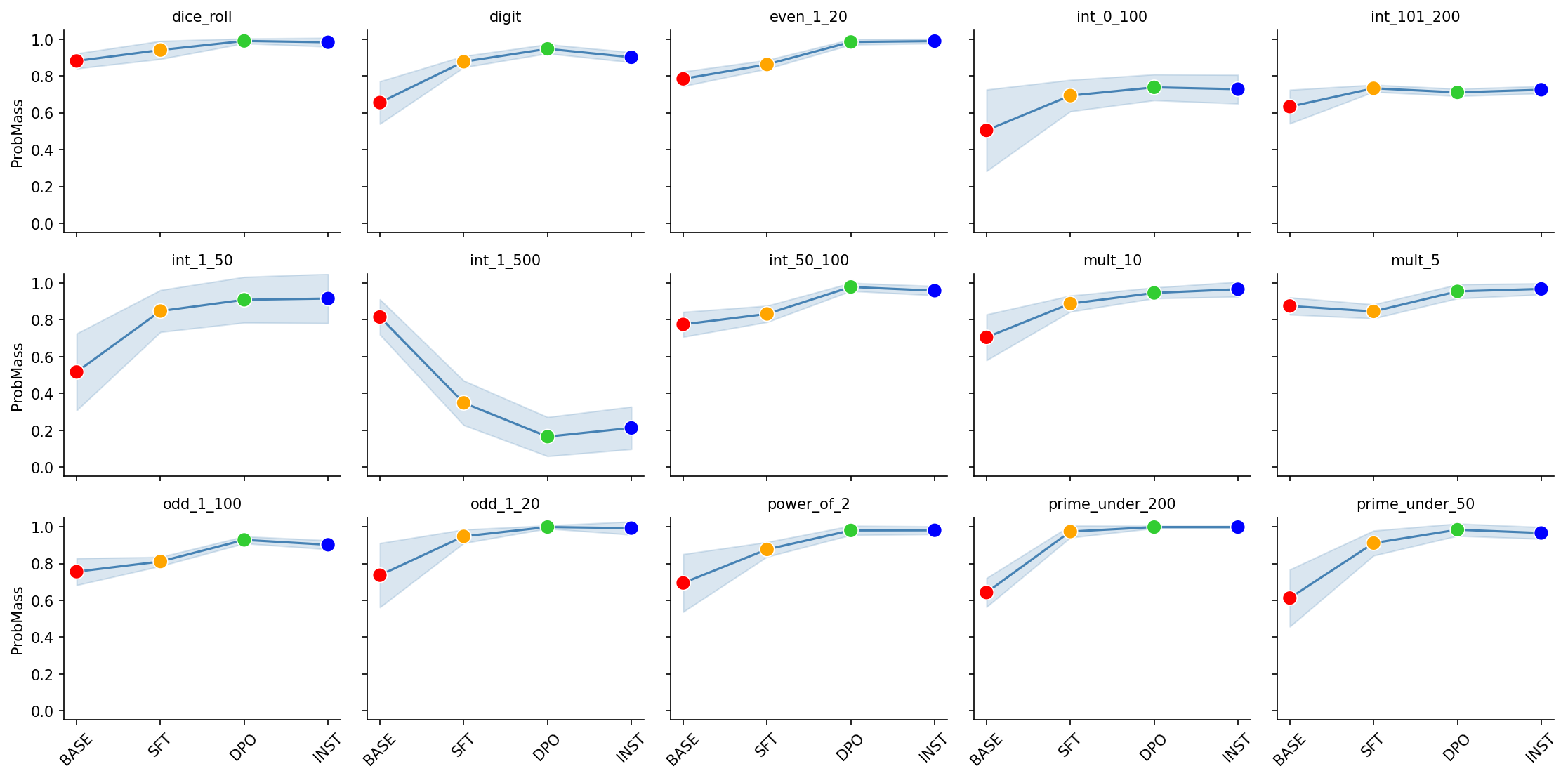}
    \caption{Probability Concentration (ProbMass)}
\end{subfigure}

\caption{Crystallization metrics for Tulu 3}
\label{fig:tulu_all_metrics}
\end{figure*}

\begin{figure*}[h]
    \includegraphics[width=\linewidth]{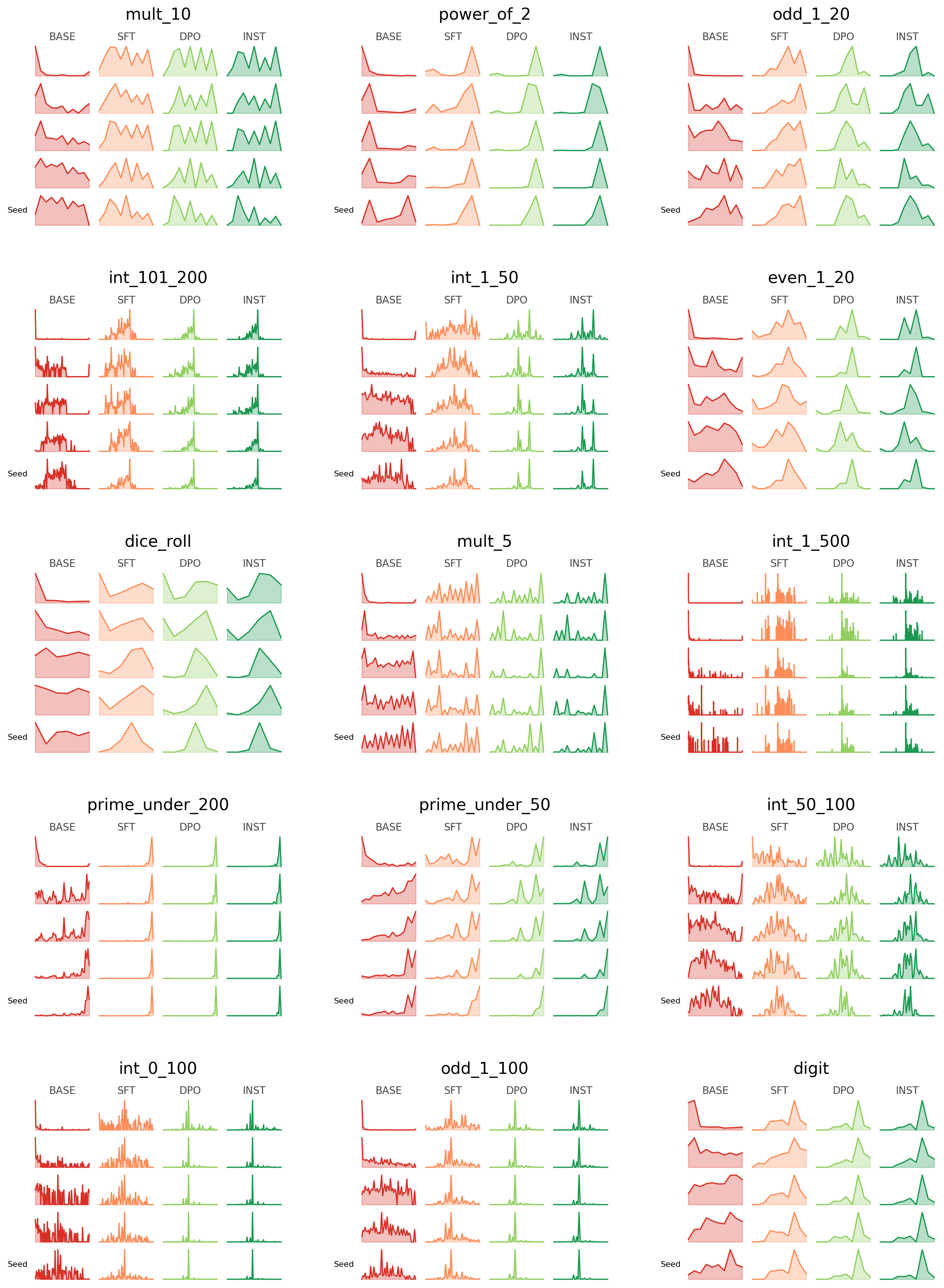}
    \caption{Task Histograms (OLMo 2)}
    \label{app:olmo_histograms}
\end{figure*}

\begin{figure*}[h]
    \includegraphics[width=\linewidth]{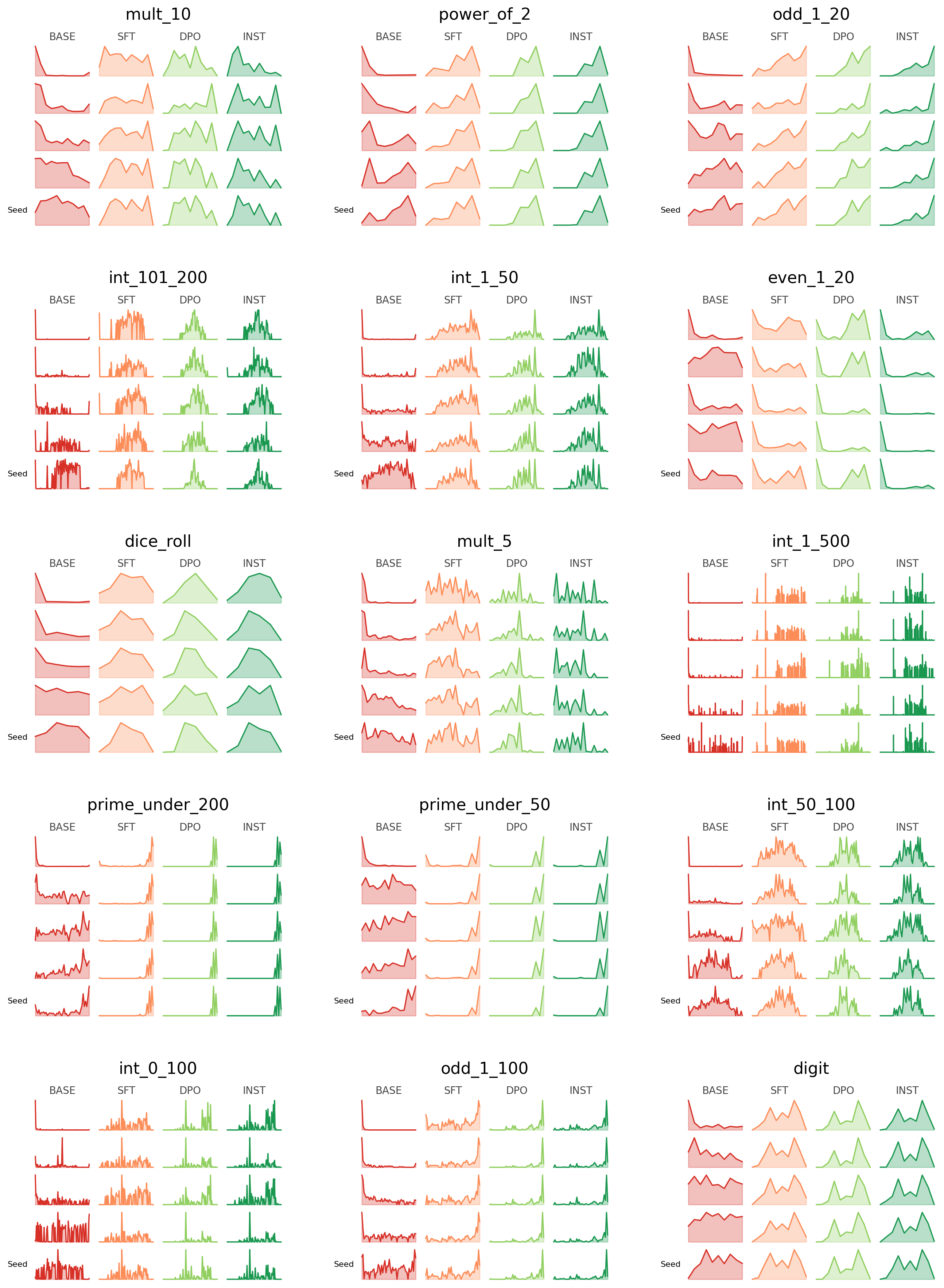}
    \caption{Task Histograms (Tulu 3)}
    \label{app:tulu_histograms}
\end{figure*}

\end{document}